%% file: main.tex
\documentclass{article}
\usepackage[preprint]{neurips_2026}
\usepackage[T1]{fontenc}
\usepackage[utf8]{inputenc}
\usepackage{microtype}
\usepackage{graphicx}
\graphicspath{{figs/}}
\usepackage{booktabs}
\usepackage{amsmath}
\usepackage{amssymb}
\usepackage{xcolor}
\usepackage{hyperref}
\usepackage{caption}
\definecolor{reclink}{RGB}{0,72,133}
\hypersetup{colorlinks=true, linkcolor=reclink, citecolor=reclink, urlcolor=reclink}
\newcommand{\arxivref}[1]{\href{https://arxiv.org/abs/#1}{arXiv:#1}}
\newcommand{\doiref}[1]{\href{https://doi.org/#1}{doi:\nolinkurl{#1}}}
\input{numbers.tex}
\newcommand{\pp}{\,pp}
\newcommand{\ci}[2]{[#1,\,#2]}
\newcommand{\C}{\textsc{C}}
\newcommand{\Sarm}{\textsc{S}}
\newcommand{\A}{\textsc{A}}
\newcommand{\B}{\textsc{B}}

\title{Reasoning Again with a Quarter of the Experts:\\ Retry First, Then Train}

\author{%
  Anik Jha\\
  Independent Researcher\\
  \texttt{anik.k.jha@gmail.com}%
}

\begin{document}
\maketitle

\begin{abstract}
Removing three quarters of the routed experts from a 122B-parameter reasoning mixture-of-experts
model makes it fit in a 12\,GiB file, and breaks it in a specific way: on LiveCodeBench it
enters verbatim reasoning loops that run to the token limit on \LoopCCappedLoop{} of
\LoopN{} development problems, where the unpruned model at the same 2-bit quantization
loops on none. We study how much of that loss can be recovered without changing the kept
experts' quantized bytes, in three steps of increasing cost. On development problems, a sampling penalty
recovers part of it. Stopping a reply when it starts to loop and drawing again recovers more on
top of it, with no weight change. Our method, \B{}, adds rank-8 trained corrections to the kept experts, loaded as an
adapter, and restores two experts per layer; it is fitted layer by layer to the unpruned
model's expert block on the pruned model's own activations. In a pre-registered comparison
on \FinalN{} easy and medium LiveCodeBench problems that no part of the study had touched, with both arms given the same
sampling penalty and the same stop-and-redraw policy, \B{} solves \FinalBSRPass{} problems
against \FinalSRPass{} (\FinalDiff\pp, 95\% CI \ci{\FinalLo}{\FinalHi}; \FinalGained{}
problems gained, \FinalLost{} lost). A training-free weight edit aimed at the loops, tested
the same way earlier, did not help. Here \B{} removes more wrong answers than loops and costs
\FinalWallRatio$\times$ the wall time.
\ifguardsdone On two older benchmarks used as regression checks, \B{} stays inside a
pre-set 2\pp{} tolerance but shows no gain (HumanEval+ \GuardHEDiff\pp, MBPP+
\GuardMBDiff\pp). \fi Refitted with the same recipe, \B{} also leads on two more MoEs at a quarter of their experts and 4 bits:
Qwen3.8-Flash-Next (\FNFinalBSRPass{} against \FNFinalSRPass, \FNFinalDiff\pp) and Laguna-S-2.1 from another
developer (\LGFinalBSRPass{} against \LGFinalSRPass, \LGFinalDiff\pp{} \ci{\LGFinalLo}{\LGFinalHi}), where it closes
a quarter of the gap to the unpruned model's \LGRefUPass. The evidence is three models, one pruning ratio and one
benchmark family; we state what it does and does not show.
\end{abstract}

\section{Introduction}

Expert pruning removes whole experts from a mixture-of-experts (MoE) language model so
that it fits in less memory. At half the experts, one-shot pruning can be close to lossless
on code generation \citep{lasby2025reap}. Deeper pruning is where memory tiers change, and
where models break. This paper is about one such break and what repairs it.

We prune Ornith-1.5-122B, an open reasoning MoE with 256 routed experts per layer, to 64
experts per layer with the REAP criterion \citep{lasby2025reap}, by copying the kept
experts' bytes out of a public 2-bit quantized file. The pruned file is \FileCGiB\,GiB. We
call this model \C{}. On LiveCodeBench \citep{jain2024livecodebench} with a reasoning
budget, \C{} does not merely get more answers wrong. It stops finishing: its reasoning
falls into verbatim repetition that runs until the token limit. Section~\ref{sec:loops}
shows that, for this model, pruning causes this and 2-bit quantization does not.

That observation frames the question. If the dominant failure is a loop, the cheapest
repairs are at decoding time and need no training at all. A repair that changes weights
has to beat those, not just beat the broken model. We therefore compare, in order of cost:
a sampling penalty (\Sarm{}); stopping a looping reply and drawing again (retry); a
training-free weight edit against the loop; and trained corrections to the expert weights
(\A{}, and \B{}, which also restores two experts per layer).

None of the ingredients is new by itself. Reconstruction-based updates of the retained
experts already improve LiveCodeBench after expert pruning down to 37.5\% retention
\citep{markovic2026optep}; low-rank matrices can compensate a compressed model while its
weights stay unchanged \citep{liu2024eora}; loops after aggressive compression of
reasoning models, and decoding-time or training-time rescues for them, have been reported
for quantization \citep{alimaskina2026lowbit,chen2026opd}. What this paper adds is a
harder regime (25\% retention on code, below the 37.5\% where that prior work stops, at
2 bits), a
controlled attribution of the loops, a weight-free baseline that a weight method must beat,
and a confirmation whose rules were fixed before the test problems were touched, repeated on two more models.

Our contributions:
\begin{itemize}
\item \textbf{A controlled attribution of the loops to pruning.} With the kept experts
byte-identical between the pruned and unpruned 2-bit files, the pruned model hits the
token limit in a verbatim loop on \LoopCCappedLoop{} of \LoopN{} problems and the
unpruned one on \LoopUtwoCappedLoop{} (Section~\ref{sec:loops}). On the third model the pruned
model's draws on the final problems end in a loop \LGFinalSRLoopDraws{} times out of \LGFinalSRDraws, and the
unpruned model's never (Section~\ref{sec:second}).
\item \textbf{Stop-and-redraw as a strong weight-free baseline.} On untouched problems,
redrawing lifts the penalised pruned model from \FinalSRWithinOne{} to \FinalSRPass{} of
\FinalN{} solved (Section~\ref{sec:results}).
\item \textbf{A weight recovery method that leaves the quantized model untouched.} \B{}
builds on the layer-wise expert weight update of \citet{markovic2026optep}: it is fitted
layer by layer against the unpruned model's expert block, but as low-rank corrections plus
two restored experts per layer, and is deployed as the pruned file plus two expert slices
per tensor plus a load-time adapter (Section~\ref{sec:method}).
\item \textbf{A pre-registered confirmation.} The finalist, the decoding policy, the
problems and the decision rule were frozen before any request on the confirmation or final
problems. \B{} leads by \FinalDiff\pp{} with a 95\% interval of \ci{\FinalLo}{\FinalHi}
(Section~\ref{sec:results}).
\item \textbf{Replications on two more models.} With the recipe, decoding policy, problems and
decision rule unchanged, and frozen again before any request on those problems, \B{} at 25\% retention solves
\FNFinalBSRPass{} of \FNFinalN{} final problems against \FNFinalSRPass{} on Qwen3.8-Flash-Next (\FNFinalDiff\pp,
\ci{\FNFinalLo}{\FNFinalHi}) and \LGFinalBSRPass{} against \LGFinalSRPass{} on Laguna-S-2.1, a model from another
family with its own architecture (\LGFinalDiff\pp, \ci{\LGFinalLo}{\LGFinalHi}). On Laguna-S-2.1 the unpruned model solves
\LGRefUPass, so \B{} closes \LGRefShare\% of the pruning gap \ci{\LGRefShareLo}{\LGRefShareHi}
(Section~\ref{sec:second}).
\item \textbf{Two negative results.} A training-free weight edit against the loops did not
transfer from development to test problems, and the sampling penalty alone was not
separable from no penalty on that test (Section~\ref{sec:decoding}).
\end{itemize}

We are explicit about scope. Everything here is three models at one pruning ratio, each in one quantization,
evaluated on one benchmark family, with one fit of each method; the attribution of the loops to pruning is for the
first and third models only. We do not
separate the contribution of the low-rank corrections from that of the restored experts on
untouched problems, and we do not explain why \B{} helps. Section~\ref{sec:limits} lists
these limits in full.

\begin{figure}[t]
\centering
\includegraphics[width=0.92\linewidth]{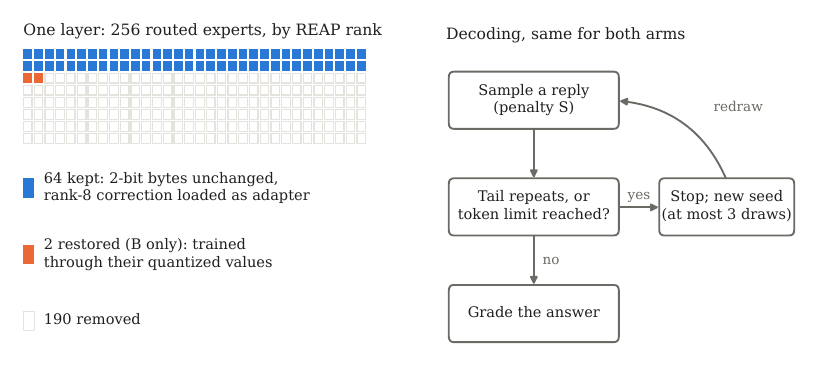}
\caption{What is compared. Left: one layer's routed experts. The pruned model \C{} keeps 64 of 256 with
their quantized bytes unchanged; \A{} and \B{} add a rank-8 correction to each kept expert as a
load-time adapter, and \B{} also restores two experts. Right: the decoding policy given to both arms
in the confirmation.}
\label{fig:overview}
\end{figure}

\section{Setting}
\label{sec:setting}

\paragraph{Model and pruning.} Ornith-1.5-122B\footnote{\url{https://huggingface.co/ornith-ai/Ornith-1.5-122B};
quantized files from \url{https://huggingface.co/unsloth/Ornith-1.5-122B-GGUF}.} has 48
layers, hidden width 3072, and in each layer 256 routed experts of width 1024, of which the
router selects 8 per token, plus one shared expert. We score experts with REAP
\citep{lasby2025reap} on our calibration data and keep the top 64 per layer. The
pruned model \C{} is built by copying the kept experts' slices and router rows out of the
public 2-bit file (UD-IQ2\_XXS), so no kept weight is requantized. The unpruned teacher
used for fitting is the 4-bit file (UD-Q4\_K\_XL) of the same model.

\paragraph{Serving.} All generation uses llama.cpp's server\footnote{\url{https://github.com/ggml-org/llama.cpp}, built from commit \texttt{36cbf415} with one local patch to chat-output parsing that is off by default.}
on one machine with 119\,GB of unified memory, with 32 parallel slots, temperature 0.6,
top-$p$ 0.95, top-$k$ 20 and a limit of 32{,}768 generated tokens unless stated. Peak
memory in the final comparison was \PeakGiB\,GiB.

\paragraph{Second model.} Qwen3.8-Flash-Next\footnote{\url{https://huggingface.co/Qwen/Qwen3.8-Flash-Next};
quantized files from \url{https://huggingface.co/unsloth/Qwen3.8-Flash-Next-GGUF}.} has 48 layers, 36 of them
recurrent and 12 with full attention, hidden width 2560, and in each layer 512 routed experts of width 640, of
which the router selects 10 per token, plus one shared expert. We prune it the same way to 128 experts per layer,
by copying the kept experts out of the public 4-bit file (UD-IQ4\_XS), which takes the file from \FNFileUGiB{} to
\FNFileCGiB\,GiB. The teacher for fitting is that same unpruned 4-bit file, so on this model the kept experts are
byte-identical to the teacher's. No Hugging Face implementation of this architecture was available to us, so the
fit runs on our PyTorch port of llama.cpp's forward pass. We accepted the port on perplexity parity with the server:
the log ratio of its perplexity to the server's is \FNParityShortLog{} at 1{,}024 tokens and \FNParityLongLog{} at 8{,}192, tolerance \FNParityShortTol. It is
served with 16 parallel slots instead of 32 to stay inside memory; peak memory in its final comparison was
\FNPeakGiB\,GiB.

\paragraph{Third model.} Laguna-S-2.1\footnote{\url{https://huggingface.co/poolside/Laguna-S-2.1}; quantized files
from \url{https://huggingface.co/unsloth/Laguna-S-2.1-GGUF}.} is from Poolside's Laguna series. Unlike the first
two models, whose files declare Qwen architectures, it has its own architecture and tokenizer. It has \LGLayers{}
layers and hidden width 3072. The first layer is dense. Each of the others has 256 routed experts of width 1024, of
which the router selects 10 per token, plus one shared expert. \LGFullAttnLayers{} layers attend to
the full context and the rest within a \LGSlidingWindow-token window. We prune it the same way, to 64 experts per
layer, by copying the kept experts out of the public 4-bit file (UD-IQ4\_XS). That takes the file from
\LGFileUGiB{} to \LGFileCGiB\,GiB. The unpruned 4-bit file is also the teacher, so the kept experts are
byte-identical to the teacher's. The fit runs on the installed Hugging Face implementation of this architecture,
loaded with the GGUF's values. We accepted it on perplexity parity with the server: the log ratios of its perplexity to the
server's are \LGParityUShortLog{}
and \LGParityULongLog{} for the unpruned file at 1{,}024 and 8{,}192 tokens, and \LGParityCShortLog{} and
\LGParityCLongLog{} for the pruned one, tolerance \LGParityTol. It is served with 16 parallel slots, a count
measured to fit before any request. Peak memory was \LGPeakGiB\,GiB in its final comparison and \LGRefPeakGiB\,GiB
for the unpruned reference run.

\paragraph{Benchmark and problem sets.} We use LiveCodeBench \citep{jain2024livecodebench}
with its official prompt and grader, restricted to easy and medium problems after a pilot
showed hard problems would exceed our compute. Three disjoint sets matter:
\begin{itemize}
\item \emph{Development}: \DevN{} problems, the earlier half by contest date of releases
v4 to v6. Every choice of method and decoding setting for the first model was made here.
\item \emph{Old test half}: \WN{} problems, the later half of v4 to v6. It was used once
to test the weight edit of Section~\ref{sec:decoding}; replies of \C{} on it were later
used as fitting data for \A{} and \B{}, so it is not a test set for them.
\item \emph{Fresh panels}: problems from releases v1 to v3, set aside before any
generation on them: \ConfN{} for confirmation and \FinalN{} for the final comparison, in
disjoint contests. We excluded problems sharing a title or contest with v4 to v6 and
checked statement overlap against every local fitting, calibration and generation file.
\end{itemize}
A problem counts as solved when the code extracted from the answer passes all hidden tests.

\paragraph{Statistics.} Comparisons are paired by problem. We report the difference in
solved fraction in percentage points (pp) with a 95\% bootstrap interval over problems
(10{,}000 resamples; 20{,}000 in Section~\ref{sec:loops}), and the exact McNemar $p$ on discordant pairs as support. Where a
problem is sampled twice, its two outcomes are averaged before resampling, so two seeds
never count as two problems.

\section{What breaks: reasoning loops}
\label{sec:loops}

Table~\ref{tab:loops} compares \C{} with the unpruned model at the same 2-bit quantization
and at 4 bits, on the development problems with a 16{,}384-token limit. The kept experts
of \C{} are byte-identical to the corresponding experts of the 2-bit unpruned file, so the
first two rows differ only in the 192 removed experts per layer.

\begin{table}[h]
\centering
\caption{Development problems (n = \LoopN), 16K token limit. ``Limit, looping'' counts
replies that reached the limit inside a verbatim token loop.}
\label{tab:loops}
\small
\begin{tabular}{lccc}
\toprule
Model & Solved & Hit the limit & Limit, looping \\
\midrule
Pruned to 25\%, 2-bit (\C) & \LoopCPass & \LoopCCapped & \LoopCCappedLoop \\
Unpruned, 2-bit & \LoopUtwoPass & \LoopUtwoCapped & \LoopUtwoCappedLoop \\
Unpruned, 4-bit & \LoopUfourPass & \LoopUfourCapped & \LoopUfourCappedLoop \\
\bottomrule
\end{tabular}
\end{table}

Pruning raises the share of problems that end in a loop at the limit by \LoopPruneDiff\pp{}
\ci{\LoopPruneLo}{\LoopPruneHi}, and accounts for a \LoopPassUtwoCDiff\pp{} drop in
solved problems \ci{\LoopPassUtwoCLo}{\LoopPassUtwoCHi}. Going from 4 bits to 2 bits
without pruning adds no loops on these problems. This scopes two earlier observations.
\citet{alimaskina2026lowbit} report that 2-bit quantization of dense 8B and 32B reasoning
models produces repetitive loops and budget exhaustion; at this scale and with this
quantizer we see none from quantization alone. \citet{song2026rethinking} report more
truncation and, under severe pruning, more repetition for expert-pruned reasoning MoEs on
LiveCodeBench at an 8K limit. They vary temperature, output caps and repetition penalties,
but have no matched quantization control and no recovery comparison of the kind studied
here; our comparison isolates pruning and the rest of this paper is about remedies. Practitioners have also
noted the effect: one model card for a REAP-pruned model reports roughly twice the loop
rate of its unpruned source and recovers it with sampler settings.\footnote{\url{https://huggingface.co/0xSero/GLM-5.2-504B},
read 2026-11-04.} Concurrent work reports runaway generation among the failures of
REAP-pruned models on production tasks and finds that the rate of repetition loops tracks
quality \citep{rasymas2026prune}, and a benchmark of stacked compression finds expert
pruning, not quantization, the dominant source of degradation
\citep{benazir2026moexbench}.

No reply of \C{} that reached the limit solved its problem
(\WTestCCappedPassed{} of \WTestCCapped{} on the old test half at 32K). A loop is a failure.

\section{Remedies that do not train weights}
\label{sec:decoding}

\paragraph{A sampling penalty (\Sarm).} The model's vendor recommends no presence penalty
for precise coding. We found that a presence penalty of 1.5 over a 20{,}480-token window
helps the pruned model: on development problems at 16K it solves \PracSPass{} against
\PracCPass{} without it, and loops at the limit drop from \PracCCapped{} to \PracSCapped.
The window matters because loop periods are long: the server's default penalty window is 64
tokens, and many repeated blocks are longer. We fix this setting as \Sarm{} for the rest of the paper.

\paragraph{A training-free weight edit.} Motivated by work showing that single-neuron
edits can remove some repetition loops \citep{lazaridis2026oneneuron}, we tested an
edit, built on the development problems, that projects out, from four attention heads, the
directions those heads write most on looping tokens. The edit was designed and chosen by a large language model (Claude Opus 5.5) acting as
editor, with access to the pruned model's weights and to development-problem activations
and results only. It is loaded as an adapter, so the
quantized file is unchanged. On development problems it solved \PracWPass{} against
\PracCPass{} for \C{}. We then froze the edit and tested it once on the old test half
at 32K (Table~\ref{tab:w}).

\begin{table}[h]
\centering
\caption{Old test half (n = \WN), 32K limit, one frozen test.}
\label{tab:w}
\small
\begin{tabular}{lcc}
\toprule
Arm & Solved & Hit the limit \\
\midrule
\C{} & \WTestCPass & \WTestCCapped \\
\C{} + weight edit & \WTestWPass & \WTestWCapped \\
\C{} + \Sarm & \WTestSPass & \WTestSCapped \\
\bottomrule
\end{tabular}
\end{table}

The edit did not transfer: \WOneDiff\pp{} against \C{} \ci{\WOneLo}{\WOneHi}. The
penalty \Sarm{} solved more than \C{} here too, but the difference, \WSminusCDiff\pp{}
\ci{\WSminusCLo}{\WSminusCHi}, is not significant on this test. We keep \Sarm{} as the
decoding setting because it is free and its direction was consistent, not because this test
established it.

\paragraph{Stop on loop, then redraw (retry).} Sampling is random, so a problem that loops
on one draw may not loop on the next. A client-side rule stops a reply as soon as its tail
repeats: for some period $P \le 6000$ characters, the last $\max(5P, 200)$ characters equal
the same-length stretch ending $P$ characters earlier, checked every 256 characters. A reply stopped by the rule,
or cut at the token limit, is drawn again with a new seed, for at most three draws per
problem. The rule was chosen on saved development replies by a criterion written before
its grid was run: it stops \TuneCappedStopped{} of \TuneCapped{} replies that had run to
the limit, at a median of \TuneStopMedian{} tokens, and would have stopped
\TuneFinishedPassedStopped{} of 360 replies that went on to finish and pass. Related work
detects repetition in reasoning traces from hidden states and continues after a cut
\citep{xie2025wordsalad}, predicts loops early from hidden states
\citep{duan2026circular}, or, for 2-bit models, detects repetitive traces and then commits
to an earlier answer or falls back to a full-precision model \citep{alimaskina2026lowbit}.
Our rule needs only the text stream and resamples from the same model.

Retry changes no weights and costs only the tokens of the abandoned draws. It is the
baseline a weight method must beat.

\section{Weight recovery}
\label{sec:method}

Both methods leave every quantized byte of \C{} in place. They add a low-rank correction
to each kept expert, applied at load time as a LoRA-style adapter \citep{hu2021lora};
\B{} also adds experts.

\paragraph{Method \A: corrections to the kept experts.} \A{} adapts the expert weight
update (EWU) of \citet{markovic2026optep}, which refits the retained experts' weights
layer by layer against the unpruned layer's output, on inputs from the already-updated
pruned model, with a regulariser toward the pretrained weights. We keep that structure and
change the parameterisation: the update is low-rank and stays outside the quantized file,
in the spirit of compensating a compressed model with low-rank matrices while its weights
stay fixed \citep{liu2024eora}, here trained per expert instead of computed in closed form.
For each layer and each kept
expert $e$, each of its three matrices $W \in \{W^{\text{gate}}_e, W^{\text{up}}_e,
W^{\text{down}}_e\}$ becomes $W + B_W A_W$ with rank 8. Layers are fitted in order. For
layer $\ell$, let $x_t$ be the input to the expert block at position $t$ when the pruned
model, with the corrections of layers $0..\ell-1$ already applied, reads the fitting
documents. Let $y_t$ be what the unpruned model's routed experts produce on the same input
$x_t$, $f_t$ the corrected pruned experts' output under the pruned model's own routing,
and $f^0_t$ their output before correction. We minimise
\begin{equation}
\mathcal{L}_\ell \;=\; \frac{\sum_t w_t \lVert y_t - f_t \rVert^2 \;+\; \mu \sum_t w_t \lVert f_t - f^0_t \rVert^2}
{\sum_t w_t \lVert y_t - f^0_t \rVert^2},
\label{eq:loss}
\end{equation}
with $\mu = 0.01$ and weights $w_t$ that give every document the same total weight. The
denominator is the layer's starting error, so the loss starts at 1. The router, attention,
shared expert and embeddings stay frozen. $B_W$ starts at zero, so training starts exactly
at \C. We use Adam with learning rate $10^{-4}$, gradient clipping at 1, and 25 passes over
the fitting documents in batches of 64 documents, and keep the checkpoint (every 5 passes)
with the lowest loss on held-out documents. No coding score is consulted during fitting.

\paragraph{Fitting data.} The documents are 1{,}018 calibration documents (256 more held
out), truncated to 2{,}048 tokens with 256 sampled positions each, plus \C{}'s own long
replies on the old test half (124 replies, each truncated to 12{,}288 tokens with 256
sampled positions, 24 of them held out for validation), so that the fit sees the hidden states the model is in when
it reasons at length; reconstructing on a model's own chain-of-thought traces was
introduced for pruning dense reasoning models by \citet{lucas2025rac}. Neither the development problems nor the fresh panels are used.

\paragraph{Method \B: \A{} plus two restored experts per layer.} \B{} starts a separate
fit from \C{} with the next two experts of each layer's REAP ranking restored from the
unpruned 2-bit file, together with their original router rows, so each layer has 66
experts and the router can select them in place of kept ones. The restored experts train
in full precision with a straight-through estimator to their deployed quantized values.
The quantized forward values are refreshed once per pass over the data, not per step, and
every validation score used for checkpoint selection quantizes afresh;
the 64 kept experts receive rank-8 corrections as in \A. Everything else follows \A,
including the data, order and token budget. Restoring two experts raises routed retention
from 25\% to 25.8\% and the file from \FileCGiB{} to \FileBGiB\,GiB; the adapter adds
\AdapterBGiB\,GiB.

\begin{figure}[t]
\centering
\includegraphics[width=0.8\linewidth]{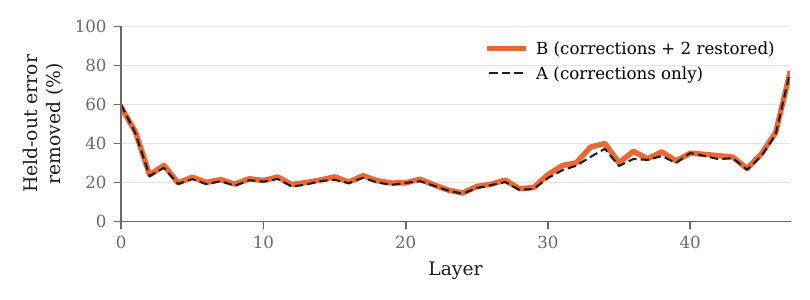}
\caption{Share of each layer's held-out reconstruction error removed by the fit.}
\label{fig:fit}
\end{figure}

\paragraph{What the fits achieve.} Across the 48 layers, the held-out reconstruction error
falls by a median of \FitARedMedian\% for \A{} (range \FitARedMin\% to \FitARedMax\%) and
\FitBRedMedian\% for \B{} (\FitBRedMin\% to \FitBRedMax\%). These are modest reductions:
most of what the removed experts computed is not recovered. Figure~\ref{fig:fit} shows that the
reduction is largest in the first and last layers and that the two fits are nearly
indistinguishable by this measure. Fitting took \FitAHours{} and
\FitBHours{} hours summed over layers on the one machine, not counting the passes that
compute each layer's inputs.

\paragraph{Deployment checks.} Before any coding evaluation we verified that the adapter
applied by the server reproduces the fitted corrections computed in our own forward pass,
that an adapter at scale zero leaves outputs byte-identical, and that every byte of \C{}
is unchanged inside the 66-expert file. Parity tolerances that proved tighter than the server's own numerical differences were
amended in dated notes before any coding result of the affected method existed;
Appendix~\ref{app:prereg} lists them.

\paragraph{Relation to prior recovery methods.} The closest method is OptEP's EWU
\citep{markovic2026optep}, described above. It updates the retained experts' full weight
matrices and is evaluated on LiveCodeBench at 37.5\% to 75\% retention; it does not
restore experts, and it reports no quantized-deployment experiment. Our changes are the low-rank, load-time form
that leaves 2-bit weights untouched, the restored experts trained through their quantized
values, fitting data that includes the pruned model's own long reasoning, and the
evaluation against a retry baseline on untouched problems. Earlier, \citet{muzio2024seer} recovered accuracy after expert pruning with
regularisation-based fine-tuning, and \citet{bercovich2026puzzle} recover a
heterogeneously expert-pruned reasoning MoE with distillation followed by reinforcement
learning. Some recovery work adapts
components other than the experts: LoRA on attention with experts and router
frozen \citep{goel2026maestro}, or distilling only the router \citep{hyeon2026routerkd}.
\citet{xie2024moepruner} recover an unstructured-pruned MoE with expert-wise distillation
from the unpruned model. \citet{hao2026lightmoe} replace redundant experts with
parameter-efficient modules and train them. \citet{lu2024naee} and \citet{zhou2025dern}
select or recombine experts without training. We do not compare against any of these methods experimentally, including dense EWU.

\section{Results}
\label{sec:results}

\subsection{Development: choosing a finalist}

On the development problems with plain decoding (one draw, 32K limit), \Sarm{} alone
solves \DevSPass{} of \DevN, \A{}+\Sarm{} \DevASPass{} and \B{}+\Sarm{} \DevBSPass.
\B{}+\Sarm{} is ahead of \Sarm{} by \DevBSminusSDiff\pp{}
\ci{\DevBSminusSLo}{\DevBSminusSHi}, and \A{}+\Sarm{} by \DevASminusSDiff\pp{}
\ci{\DevASminusSLo}{\DevASminusSHi}. By a selection rule fixed in advance (a candidate
must be at least 2\pp{} above \Sarm{}; \B{} is chosen over \A{} only if at least 2\pp{}
above it), \B{} became the development winner. This is a selection, not a significance
claim, and \A{} was not carried further.

A post hoc look at these runs showed that a plain rerun of the same arm changes the
outcome on many problems, and that a rerun rescues limit-hitting problems about as often
as \B{} does. That is what prompted the retry baseline: before crediting \B{} with a
recovery, both arms should be allowed to redraw. With retry on both arms, on the same
development problems, \B{}+\Sarm{} solves \RetryBSRPass{} and \Sarm{} \RetrySRPass{}
(\RetryDiff\pp, \ci{\RetryLo}{\RetryHi}). This is still development evidence: \B{} was
selected and the loop rule tuned on these problems.

\subsection{Pre-registered confirmation on untouched problems}

We then froze, in one file with hashes of the code, both model files, the adapter and the
server build: the finalist (\B{}+\Sarm{}+retry), the baseline (\Sarm{}+retry), the
sampling settings and seeds, the fresh panels, and the decision rules. The plan had two
stages. A confirmation stage on \ConfN{} problems with two paired seeds acts only as a
gate: a positive mean difference permits the final stage. The final stage is one paired
sample on \FinalN{} further problems; \B{} ``leads'' only if the difference is at least
2\pp{} and its 95\% interval lies above zero, and an interval that includes zero is
reported as inconclusive. No arm, setting, problem or decision rule was changed after the first
request on a confirmation problem; one later amendment raised the compute ceiling and was not
needed (Appendix~\ref{app:prereg}).

\begin{table}[h]
\centering
\caption{\Sarm{}+retry against \B{}+\Sarm{}+retry: problems solved after at most three
draws. Differences in percentage points with 95\% bootstrap intervals over problems. All
units ran to completion with no request errors.}
\label{tab:main}
\small
\setlength{\tabcolsep}{4.5pt}
\begin{tabular}{lcccccc}
\toprule
Problems & $n$ & \Sarm{}+retry & \B{}+\Sarm{}+retry & Difference & Gained / lost & McNemar $p$ \\
\midrule
Development (reused) & \DevN & \RetrySRPass & \RetryBSRPass & \RetryDiff{} \ci{\RetryLo}{\RetryHi} & \RetryGained{} / \RetryLost & \RetryP \\
Confirmation, seed 1 & \ConfN & \ConfSRonePass & \ConfBSRonePass & \ConfOneDiff{} \ci{\ConfOneLo}{\ConfOneHi} & \ConfOneGained{} / \ConfOneLost & \ConfOneP \\
Confirmation, seed 2 & \ConfN & \ConfSRtwoPass & \ConfBSRtwoPass & \ConfTwoDiff{} \ci{\ConfTwoLo}{\ConfTwoHi} & \ConfTwoGained{} / \ConfTwoLost & \ConfTwoP \\
Confirmation, pooled & \ConfN & & & \ConfPooledDiff{} \ci{\ConfPooledLo}{\ConfPooledHi} & & \\
\textbf{Final (untouched)} & \FinalN & \textbf{\FinalSRPass} & \textbf{\FinalBSRPass} & \textbf{\FinalDiff{} \ci{\FinalLo}{\FinalHi}} & \FinalGained{} / \FinalLost & \FinalP \\
\bottomrule
\end{tabular}
\end{table}

\begin{figure}[t]
\centering
\includegraphics[width=0.78\linewidth]{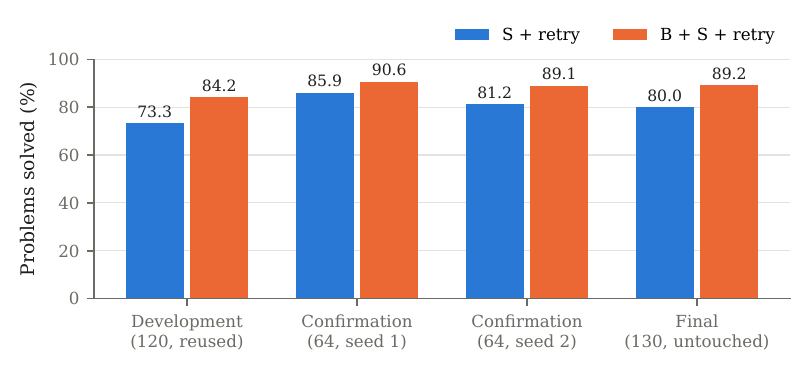}
\caption{Problems solved after at most three draws, by problem set. The final set is the
one pre-registered primary comparison.}
\label{fig:stages}
\end{figure}

Table~\ref{tab:main} and Figure~\ref{fig:stages} give the outcome. The confirmation gate
passed with a pooled difference of \ConfPooledDiff\pp. On the final problems \B{} solves
\FinalBSRPass{} (\FinalBSRPassPct\%) against \FinalSRPass{} (\FinalSRPassPct\%), a
difference of \FinalDiff\pp{} with interval \ci{\FinalLo}{\FinalHi}, and the frozen rule
returns ``\FinalDecision''. We recomputed the counts and the interval from the raw result
rows. The effect is smaller on the confirmation problems than on the development
problems, as expected for a method selected on the latter, and neither confirmation seed
alone excludes zero.

The original plan compared the two arms without retry. As a secondary, the first draws
alone give a first-draw comparison with loop stopping (a reply stopped for looping counts
as a failure, so this is not the plan's plain comparison exactly): \FinalFirstDrawDiff\pp{}
\ci{\FinalFirstDrawLo}{\FinalFirstDrawHi}.

\subsection{Where the gain is, and what it costs}

\begin{figure}[t]
\centering
\includegraphics[width=0.86\linewidth]{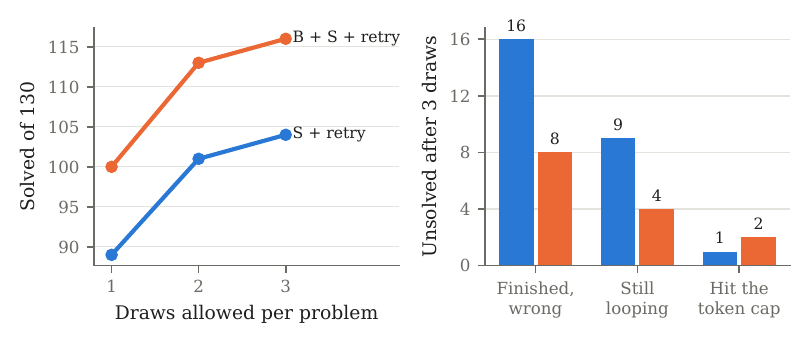}
\caption{Final problems (n = \FinalN). Left: problems solved when one, two or three draws
are allowed. Right: why the unsolved problems failed after three draws.}
\label{fig:breakdown}
\end{figure}

\paragraph{Redrawing helps both arms.} Figure~\ref{fig:breakdown} (left) shows that the
baseline goes from \FinalSRWithinOne{} solved on the first draw to \FinalSRPass{} after
three, and \B{} from \FinalBSRWithinOne{} to \FinalBSRPass. Redrawing is worth more to
either arm than the gap between the arms, which is why a weight method should be compared
against a baseline that is also allowed to redraw. The gap itself is about the same on the
first draw as after three.

\paragraph{\B{} removes more wrong answers than loops.} After three draws the baseline
fails \FinalSRFailWrong{} problems with a finished wrong answer and \FinalSRFailLoop{}
still looping; \B{} fails \FinalBSRFailWrong{} and \FinalBSRFailLoop{}
(Figure~\ref{fig:breakdown}, right). The gain is concentrated on medium problems
(\FinalSRMedium{} to \FinalBSRMedium{} of \FinalMediumN{}; easy \FinalSREasy{} to
\FinalBSREasy{} of \FinalEasyN). We set out to repair loops and found that the weight
method's contribution lies largely elsewhere. We report this as an observation; we have
not measured why.

\paragraph{Cost.} \B{} took \FinalBSRWall{} hours of wall time on the final problems
against \FinalSRWall{} for the baseline (\FinalWallRatio$\times$), with
\FinalBSRTokens{}K against \FinalSRTokens{}K generated tokens over \FinalBSRDraws{} and
\FinalSRDraws{} draws. We did not decompose the slowdown. The arms are compared at equal draws, not equal time.

\paragraph{Regression guards.} The plan requires that \B{} lose no more than 2\pp{} to the
identically decoded baseline on HumanEval+ and MBPP+ \citep{liu2023evalplus} before it is
adopted.
\ifguardsdone
On HumanEval+ \B{} solves \GuardHEBSR{} of \GuardHEN{} against \GuardHESR{}
(\GuardHEDiff\pp, \ci{\GuardHELo}{\GuardHEHi}). On MBPP+ it solves fewer: \GuardMBBSR{} of
\GuardMBN{} against \GuardMBSR{} (\GuardMBDiff\pp, \ci{\GuardMBLo}{\GuardMBHi};
\GuardMBGained{} gained, \GuardMBLost{} lost). The frozen rule compares the point difference
with the tolerance, and by it the guards \GuardVerdict, so \B{} meets the plan's adoption
rule. The MBPP+ margin is the narrowest
possible: one more lost problem would have failed the rule, and the interval does not
exclude a loss larger than the tolerance. We therefore read MBPP+ as no evidence of a gain
on short problems and weak evidence against a large loss, and we claim the gain of \B{}
for LiveCodeBench at a reasoning budget only. These benchmarks are public and
old; they are regression checks, not untouched evidence. Figure~\ref{fig:forest} puts
every paired comparison of the two arms on one axis.
\else
These runs were not complete when this version was written, so we report the
LiveCodeBench result as ``leads, guards pending'' and do not call \B{} adopted.
\fi

\ifguardsdone
\begin{figure}[t]
\centering
\includegraphics[width=0.82\linewidth]{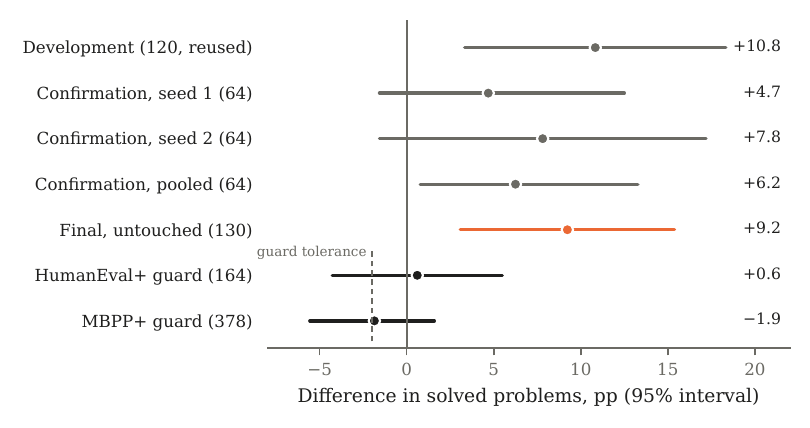}
\caption{\B{}+\Sarm{}+retry minus \Sarm{}+retry on every problem set, with 95\% bootstrap intervals.
The final set is the pre-registered primary comparison; the two guards are regression checks
with a tolerance of 2\pp.}
\label{fig:forest}
\end{figure}
\fi

\subsection{Two more models}
\label{sec:second}

\paragraph{Qwen3.8-Flash-Next.} We repeated the confirmation on Qwen3.8-Flash-Next (Section~\ref{sec:setting})
without choosing anything on its development problems. \B{}
was fitted with the recipe of Section~\ref{sec:method} unchanged: rank, learning rate, passes, fitting documents,
and the replies of this model's \C{} on the same old test half. The arms, sampling, loop rule, retry policy, panels and decision
rule are those of the first model's pre-registration, frozen again with this model's files before any request on
a fresh problem. Nothing was tuned on this model. Across its 48 layers the fit removes a median \FNFitRedMedian\%
of the held-out reconstruction error (\FNFitRedMin\% to \FNFitRedMax\%; \FitBRedMedian\% on the first model).
Restoring two experts raises routed retention from 25\% to 25.4\% and the file to \FNFileBGiB\,GiB; the adapter
adds \FNAdapterBGiB\,GiB. Two deployment-parity checks failed as first written. Both were changed after their
data had been seen, before any coding result of \B{} existed; Appendix~\ref{app:second} gives both.

This model's \C{} also loops. With one draw and no penalty, the loop rule stopped its reply on \FNDevCLoop{} of
\FNDevN{} development problems. We did not run the unpruned model on these problems, so we do not attribute
those loops to pruning.

\begin{table}[h]
\centering
\caption{The two replications at 25\% retention: \Sarm{}+retry against \B{}+\Sarm{}+retry on the first model's
confirmation and final problems, solved after at most three draws. Differences in percentage points with 95\%
bootstrap intervals over problems. All units ran to completion. On Qwen3.8-Flash-Next one timed-out request was
repeated with its seed, as the frozen plan allows; Laguna-S-2.1 had no request error. Laguna-S-2.1's unpruned model
solves \LGRefUPass{} of the \LGFinalN{} final problems under the same settings.}
\label{tab:second}
\small
\setlength{\tabcolsep}{4.5pt}
\begin{tabular}{lcccccc}
\toprule
Problems & $n$ & \Sarm{}+retry & \B{}+\Sarm{}+retry & Difference & Gained / lost & McNemar $p$ \\
\midrule
\multicolumn{7}{l}{\emph{Qwen3.8-Flash-Next}} \\
Confirmation, seed 1 & \FNConfN & \FNConfSRonePass & \FNConfBSRonePass & \FNConfOneDiff{} \ci{\FNConfOneLo}{\FNConfOneHi} & \FNConfOneGained{} / \FNConfOneLost & \FNConfOneP \\
Confirmation, seed 2 & \FNConfN & \FNConfSRtwoPass & \FNConfBSRtwoPass & \FNConfTwoDiff{} \ci{\FNConfTwoLo}{\FNConfTwoHi} & \FNConfTwoGained{} / \FNConfTwoLost & \FNConfTwoP \\
Confirmation, pooled & \FNConfN & & & \FNConfPooledDiff{} \ci{\FNConfPooledLo}{\FNConfPooledHi} & & \\
\textbf{Final} & \FNFinalN & \textbf{\FNFinalSRPass} & \textbf{\FNFinalBSRPass} & \textbf{\FNFinalDiff{} \ci{\FNFinalLo}{\FNFinalHi}} & \FNFinalGained{} / \FNFinalLost & \FNFinalP \\
\midrule
\multicolumn{7}{l}{\emph{Laguna-S-2.1}} \\
Confirmation, seed 1 & \LGConfN & \LGConfSRonePass & \LGConfBSRonePass & \LGConfOneDiff{} \ci{\LGConfOneLo}{\LGConfOneHi} & \LGConfOneGained{} / \LGConfOneLost & \LGConfOneP \\
Confirmation, seed 2 & \LGConfN & \LGConfSRtwoPass & \LGConfBSRtwoPass & \LGConfTwoDiff{} \ci{\LGConfTwoLo}{\LGConfTwoHi} & \LGConfTwoGained{} / \LGConfTwoLost & \LGConfTwoP \\
Confirmation, pooled & \LGConfN & & & \LGConfPooledDiff{} \ci{\LGConfPooledLo}{\LGConfPooledHi} & & \\
\textbf{Final} & \LGFinalN & \textbf{\LGFinalSRPass} & \textbf{\LGFinalBSRPass} & \textbf{\LGFinalDiff{} \ci{\LGFinalLo}{\LGFinalHi}} & \LGFinalGained{} / \LGFinalLost & \LGFinalP \\
\bottomrule
\end{tabular}
\end{table}

Table~\ref{tab:second} gives the outcome. The baseline is weaker on this model than on the first: it solves
\FNConfSRonePass{} and \FNConfSRtwoPass{} of \FNConfN{} confirmation problems, against \ConfSRonePass{} and
\ConfSRtwoPass. \B{} solves \FNConfBSRonePass{} and \FNConfBSRtwoPass, so the confirmation difference is larger:
\FNConfPooledDiff\pp{} pooled \ci{\FNConfPooledLo}{\FNConfPooledHi}, with no problem lost under either seed. On
the final problems \B{} solves \FNFinalBSRPass{} against \FNFinalSRPass{} (\FNFinalDiff\pp,
\ci{\FNFinalLo}{\FNFinalHi}), and the frozen rule returns ``\FNFinalDecision''. We recomputed the counts and
intervals from the raw result rows. On HumanEval+ \B{} solves \FNGuardHEBSR{} of \FNGuardHEN{}
against \FNGuardHESR{} (\FNGuardHEDiff\pp, \ci{\FNGuardHELo}{\FNGuardHEHi}), and on MBPP+ \FNGuardMBBSR{} of
\FNGuardMBN{} against \FNGuardMBSR{} (\FNGuardMBDiff\pp, \ci{\FNGuardMBLo}{\FNGuardMBHi}); by the frozen rule the
guards \FNGuardVerdict.

As on the first model, \B{} removes more finished wrong answers than loops: after three draws the baseline fails
\FNFinalSRFailWrong{} problems with a finished wrong answer and \FNFinalSRFailLoop{} still looping, \B{}
\FNFinalBSRFailWrong{} and \FNFinalBSRFailLoop. Medium problems go from \FNFinalSRMedium{} to
\FNFinalBSRMedium{} of \FNFinalMediumN, easy ones from \FNFinalSREasy{} to \FNFinalBSREasy{} of
\FNFinalEasyN. The first draws alone give \FNFinalFirstDrawDiff\pp{}
\ci{\FNFinalFirstDrawLo}{\FNFinalFirstDrawHi}. We have not measured why the gain is larger on this model.

\B{} costs more here than on the first model. It took \FNFinalWallRatio$\times$ the baseline's wall time on the
final problems (\FinalWallRatio$\times$ on the first model), with \FNFinalBSRTokens{}K against \FNFinalSRTokens{}K
generated tokens over \FNFinalBSRDraws{} and \FNFinalSRDraws{} draws. We did not decompose the slowdown.

\paragraph{Laguna-S-2.1.} This model's pruned build failed a screen we had fixed before fitting anything. With the
sampling penalty and up to three draws, \C{} solved \LGScreenCPass{} of \LGScreenN{} development problems against
the unpruned model's \LGScreenUPass, below the half of the unpruned score we had required. We ran \B{} anyway. That
decision was made after the screen's result, and every other rule stayed as written. \B{} was fitted with the
recipe of Section~\ref{sec:method} unchanged. The second model's amended deployment-parity rules applied from the
start, and the confirmation files were frozen with this model's files before any request on a fresh problem
(Appendix~\ref{app:second}). Only \LGFitTextBeforeLoop\% of the text in \C{}'s own fitting replies comes before
their first loop (\LGFitTextBeforeLoopOrnith\% on the first model, \LGFitTextBeforeLoopFlashnext\% on the second),
so most of its fitting text is loop text. Across the 47 expert layers the fit still removes a median
\LGFitRedMedian\% of the held-out reconstruction error (\LGFitRedMin\% to \LGFitRedMax\%).

Here the loops can be attributed to pruning, as on the first model. \C{}'s kept experts are byte-identical to the
unpruned file, and that file fits in memory. On the final problems, under the same penalty and retry policy,
\LGFinalSRLoopDraws{} of \C{}'s \LGFinalSRDraws{} draws ended in a loop and none of the unpruned model's
\LGRefUDraws. \C{} also opened \LGFinalSREmptyThink{} of its draws with an empty reasoning block; \B{} and the
unpruned model never did.

\B{} solves \LGConfBSRonePass{} and \LGConfBSRtwoPass{} of \LGConfN{} confirmation problems against
\LGConfSRonePass{} and \LGConfSRtwoPass{} (\LGConfPooledDiff\pp{} pooled, \ci{\LGConfPooledLo}{\LGConfPooledHi}).
On the final problems it solves \LGFinalBSRPass{} against \LGFinalSRPass{} (\LGFinalDiff\pp,
\ci{\LGFinalLo}{\LGFinalHi}; \LGFinalGained{} gained, \LGFinalLost{} lost), and the frozen rule returns
``\LGFinalDecision''. On HumanEval+ \B{} solves \LGGuardHEBSR{} of \LGGuardHEN{} against \LGGuardHESR{}
(\LGGuardHEDiff\pp, \ci{\LGGuardHELo}{\LGGuardHEHi}), and on MBPP+ \LGGuardMBBSR{} of \LGGuardMBN{} against
\LGGuardMBSR{} (\LGGuardMBDiff\pp, \ci{\LGGuardMBLo}{\LGGuardMBHi}). By the frozen rule the guards
\LGGuardVerdict. The MBPP+ interval stays above the $-$2\pp{} tolerance but includes zero, so we read it as no loss beyond the
tolerance, not as a gain.

How much of the loss does \B{} recover? After the guards, from a file frozen before its first request, we ran the
unpruned model once on the final problems with the baseline's exact settings. It solves \LGRefUPass{} of
\LGFinalN. Pruning therefore costs \LGRefGap{} solved problems, and \B{} solves \LGRefClosed{} more than \C{}:
\LGRefShare\% of the gap (95\% bootstrap interval over problems, \LGRefShareLo\% to \LGRefShareHi\%). \B{} stays
\LGRefBelowU\pp{} below the unpruned model \ci{\LGRefBelowULo}{\LGRefBelowUHi}. Most of what pruning
removed is still missing.

On this model, unlike the first two, \B{}'s gain is mostly in loops. After three draws the baseline fails
\LGFinalSRFailWrong{} problems with a finished wrong answer and \LGFinalSRFailLoop{} still looping; \B{} fails
\LGFinalBSRFailWrong{} and \LGFinalBSRFailLoop. Medium problems go from \LGFinalSRMedium{} to \LGFinalBSRMedium{}
of \LGFinalMediumN, easy ones from \LGFinalSREasy{} to \LGFinalBSREasy{} of \LGFinalEasyN. The first draws alone
give \LGFinalFirstDrawDiff\pp{} \ci{\LGFinalFirstDrawLo}{\LGFinalFirstDrawHi}. We have not measured why the pattern
differs between models.

On this model \B{} cost little extra wall time: \LGFinalWallRatio$\times$ the baseline's on the final problems
(\FinalWallRatio$\times$ on the first model, \FNFinalWallRatio$\times$ on the second), while it generated fewer
tokens, \LGFinalBSRTokens{}K against \LGFinalSRTokens{}K over \LGFinalBSRDraws{} and \LGFinalSRDraws{} draws. We
did not decompose the wall time.

\paragraph{Caveats specific to the replications.} \enlargethispage{8pt}Both reused the first model's final problems, and \B{} was carried
to each after it had led before: to Qwen3.8-Flash-Next after the first model, to Laguna-S-2.1 after both. Nothing
in \B{} or the protocol changed after those results, and no reply of either model on these problems existed before
its freeze. But the choice of which method to repeat was made with earlier results known, and on Laguna-S-2.1 the
decision to run at all came after its screen had failed.

\section{Limitations}
\label{sec:limits}

\begin{itemize}
\item \textbf{Three models, one ratio.} All three are pruned to 25\% routed retention, each in one quantization
(2-bit, 4-bit and 4-bit). Nothing was chosen on the second model's development problems, and it had no unpruned reference run, so for it
we cannot say how much of the loss pruning causes. The third model's pruned build failed our pre-set screen, and we
ran \B{} on it after seeing that result. Both later models reused the first model's test problems, and \B{} was
carried to each after it had led before (Section~\ref{sec:second}). Two of the second model's deployment-parity
rules were changed after their data had been seen (Appendix~\ref{app:second}).
\item \textbf{Recovery is partial.} On the one model where we measured it, \B{} closes \LGRefShare\% of the
gap in solved problems that pruning opened \ci{\LGRefShareLo}{\LGRefShareHi} and stays \LGRefBelowU\pp{} below the unpruned model.
\item \textbf{One fit, one decoding setting, one benchmark family.} Each method was fitted
once with one seed and recipe; no rank, data or learning-rate search was run.
\item \textbf{\B{} bundles two changes.} It adds low-rank corrections and restores two
experts per layer, which also adds 0.8 points of routed capacity (0.4 on the second model, 0.8 on the third). \A{} (corrections only)
was evaluated on development problems only, and an untrained 66-expert control was not
evaluated on code. We cannot say which component carries the gain.
\item \textbf{Equal draws, not equal cost.} \B{} is \FinalWallRatio$\times$ slower in wall
time under our serving setup (\FNFinalWallRatio$\times$ on the second model, \LGFinalWallRatio$\times$ on the third). These ratios mix
the adapter's cost per token with each arm's token count, which differ; we did not measure the adapter's cost per
token on its own.
\item \textbf{Public, older problems.} The fresh panels were untouched by this study and
screened for overlap with our own data, but they were published before the model was
released; exposure during pretraining is shared by both arms and can compress differences.
\item \textbf{A design change after development.} The pre-registered arms use retry, which
the original plan did not have; the change was made on development evidence and frozen
before any fresh-panel request. A first-draw comparison with loop stopping is reported as
a secondary; it is close to the plan's original comparison but not identical to it.
\item \textbf{The ingredients are known.} That updating retained experts against a
reconstruction target helps after pruning, and that low-rank matrices can compensate a
compressed model, are prior results (Section~\ref{sec:method}). Our evidence concerns a
more aggressive regime, a stronger baseline and a pre-registered test, not a new principle.
\item \textbf{No comparison with other recovery methods.} In particular we do not compare
with the dense expert weight update that \A{} is derived from, so we do not claim the
low-rank form is better than it, only that it leaves the quantized file unchanged. We did not
compare against activation steering \citep{nguyen2026stop}, a training-free remedy for the same
failure.
\item \textbf{No account of mechanism.} We have not measured why \B{} helps, why its gain shows on finished wrong
answers more than on loops on the first two models and the reverse on the third, or why the pruned third model
often skips its reasoning.
\item \textbf{Sampling at 32 parallel slots} (16 for the other two models) does not pin a reply to its seed, so
run-to-run variation is part of every number here; the two confirmation seeds show its
size.
\end{itemize}

\section{Related work}
\label{sec:related}

\paragraph{Expert pruning and recovery.} REAP scores experts by router gate values and
activation norms and reports near-lossless code generation at 50\% on very large MoEs
\citep{lasby2025reap}; we use it as the pruning step and study what follows at 25\%. Our
earlier work found that half the experts could be removed from two coding MoEs with no
statistically detectable loss on its primary code benchmark \citep{jha2026half}. A short full
fine-tune recovers about a third of the gap after expert pruning or merging
\citep{hyeon2026beyond}, COMPEL compensates the kept experts with a Fisher-guided update
\citep{yoon2026compel}, and low-rank compensators already exist for quantized MoEs
\citep{huang2025milo}. The
recovery methods closest to ours are discussed in Section~\ref{sec:method}.

\paragraph{Loops in reasoning models.} \citet{pipis2025wait} study why reasoning models
loop and find that students distilled from a teacher loop more than the teacher; a pruned
model is also an imperfect copy of its source, though we do not test their account.
\citet{xie2025wordsalad} detect repetitive reasoning from hidden states and continue with
a prompt. \citet{lazaridis2026oneneuron} show that editing single neurons can suppress
specific repetition failures and that the effect depends on the model; our training-free
edit is a negative data point in the same direction. \citet{alimaskina2026lowbit} and
\citet{song2026rethinking} report repetition after extreme quantization and after pruning
respectively (Section~\ref{sec:loops}). \citet{nguyen2026stop} see the same failure to stop after
weight pruning of dense reasoning models and recover part of it by steering activations;
\citet{zhang2026shortopd} find for depth-pruned dense models that pass@$k$ recovers
substantially under repeated sampling while greedy pass@1 nearly vanishes.
\citet{ettifouri2026cusum} monitor quantized small reasoning models for repetitive or
unproductive trajectories and, after an alarm, re-decode from an earlier state; our retry arm
instead restarts the reply with a new seed. \citet{chen2026opd} trace loops in sub-3-bit
reasoning models to training on fixed prefixes and reduce them by distilling on the
quantized model's own trajectories; our fitting data includes the pruned model's own
replies for a similar reason, though on the first two models \B{}'s measured gain was mostly not on loops.

\section{Conclusion}

At a quarter of its experts, this reasoning MoE fails mostly by looping, and much of that
is recoverable without training: a sampling penalty recovers part on development problems, and stopping and redrawing adds more. On top
of that baseline, a small trained correction to the kept experts plus two restored experts
per layer solved \FinalBSRPass{} of \FinalN{} untouched problems against \FinalSRPass, in
a comparison whose rules were fixed in advance. \ifguardsdone On two older, shorter
benchmarks it showed no gain, and on one a small loss inside the pre-set tolerance. \fi
The comparison came out the same way on two more models at the same retention, one of them from another family
with its own architecture. On that model the unpruned reference shows that \B{} closes about a quarter of the
pruning gap, so most of what pruning removed is still missing. The next steps are a fresh-panel test that
separates the corrections from the restored experts, a test at other retention levels, and a cheaper way to apply
the corrections at inference.

\paragraph{Artifacts.} The frozen pre-registration files, per-problem results, attempt
logs and analysis code will be released with a later version of this paper.

\appendix

\section{Pre-registration record}
\label{app:prereg}

Each stage of the first model's study that produced a number in this paper was run from a file frozen before
the stage's first result: it lists the arms, settings, seeds, problems and decision rule, and
the SHA-256 of every code file, model file, adapter and server binary the stage uses. The
driver re-checks every hash at start and refuses to run on a mismatch. A unit of work (one
arm on one problem set with one seed) counts as complete only if every problem is graded
once, no request error is outstanding, the graded replies are exactly those the attempt log
gives under the frozen retry policy, and the recorded settings equal the frozen ones.

\paragraph{Amendments.} Five dated amendments exist; none changes a coding-evaluation
rule. Two (one each for \A{} and \B{}) replaced a numerical-parity tolerance that had been
set below the server's own difference between mathematically identical settings, so that
the unmodified pruned model failed it against itself. One replaced a deployment-parity
fixture for \B{} that had narrowly failed its tolerance with a stronger fixture under the
same tolerances, fixed before the stronger fixture ran. One added a deployment-parity
check of the fitted \B{} that the original file lacked. These four were written before any
coding result of the affected method existed. The fifth raised the compute ceiling of the
final stages after the final decision had been recorded, in case the regression guards
needed it; they finished inside the original ceiling.

\paragraph{Decision rule for the final comparison.} Let $d$ be the difference in solved
fraction, finalist minus baseline, on the replies standing after at most three draws, with
a 95\% bootstrap interval over problems. \emph{Leads}: $d \ge 0.02$ and the interval lies
above zero. \emph{Inconclusive}: the interval includes zero, whatever $d$ is. \emph{Below
margin}: interval above zero and $d < 0.02$. \emph{Worse}: interval below zero.

\paragraph{Compute.} The confirmation and final stages used \GpuHours{} hours of
generation and grading on one machine.

\section{Pre-registration record for the other two models}
\label{app:second}

The second model's fit and its confirmation were each run from a file frozen before the stage's first result,
with the same hash checks as Appendix~\ref{app:prereg}. Its settings are the first model's except for the
following, all written into the frozen files. The second model's development run and the third model's screen,
both reported in Section~\ref{sec:second}, followed plans written before their first request but were not run
from hashed files.

\paragraph{Second model: settings changed, before any result.} 16 parallel slots instead of 32, chosen by a
memory measurement fixed in advance. A memory cap for the fit, set from a smoke-test measurement by a rule fixed
before it ran, because this architecture's hidden state per token is larger. \C{}'s own replies for the fitting
data were capped at 12{,}288 generated tokens, the length the fit truncates them to anyway. A tokenizer built from
the first model's, whose first \FNTokShared{} tokens and all merges are identical to this model's, plus this model's
\FNTokExtra{} extra tokens and its chat template, checked against the GGUF. The PyTorch port described in
Section~\ref{sec:setting}. The regression-guard file was frozen together with the confirmation file, not after the
final decision. The compute ceiling for the confirmation, final and guards together was \FNBudgetHours{} hours of generation and grading.

\paragraph{Second model: two gate changes written after seeing data.} Both concern deployment parity of \B{}, and
both were made before any coding result of \B{} existed. (1) A check that the 130-expert file with the two restored
experts switched off reproduces \C{} required \FNDisabledGreedyMin\% agreement of greedy next tokens. It failed at
\FNDisabledGreedy\%, and a diagnostic showed that \C{} agrees with itself under mathematically identical server
settings only \FNDisabledFloorLo\% to \FNDisabledFloorHi\% of the time. As in Appendix~\ref{app:prereg}, the
threshold was replaced by that floor, \FNDisabledFloorLo\%; every other criterion was unchanged, and the byte checks of the file passed
independently. (2) A fixture that applies the fitted corrections at every second layer failed its relative
tolerance (\FNFixtureRelRms{} against \FNFixtureTol), with a residual that did not scale with the edit
(across text chunks, correlation \FNFixtureResidualCorr{} between the residual and the edit's effect in the
server). The fixture was strengthened to every layer with the tolerances unchanged, a
rule fixed before it ran, and passed (\FNFixtureAllRelRms).

\paragraph{Second model: compute.} The second model's confirmation and final stages used \FNGpuHours{} hours of
generation and grading, and its regression guards a further \FNGuardGpuHours{} hours.

\paragraph{Third model: the decision to run.} The plan for this model said a failed screen ends the study for it.
The screen failed (\LGScreenCPass{} of \LGScreenN{} development problems against a bar of half the unpruned
model's \LGScreenUPass). The decision to continue was taken after that result, and the frozen files say so.

\paragraph{Third model: settings changed, before any result.} 16 parallel slots, by the same measured rule. \C{}'s
replies for the fitting data were capped at 12{,}288 generated tokens, as on the second model. The model's
published tokenizer and llama.cpp split a carriage return differently, so \LGTokDocsCarriageReturn{} of the
\LGTokDocs{} calibration documents tokenize differently between them; the fit reads the token ids the server reads. The
GGUF's rotary scaling (YaRN factor \LGYarnGGUF) differs from the published configuration's (\LGYarnConfig); the
fit follows the GGUF, which is what the server runs. The dense first layer is run, not fitted. The go/no-go check
before the full fit, which requires that fitting four layers lowers their held-out error, uses the first four expert layers. The compute ceiling for the confirmation, final and guards together was
\LGBudgetHours{} hours of generation and grading.

\paragraph{Third model: gates.} One check was changed after its data had been seen, before any fitting data
existed. The tokenizer check failed on the carriage-return split above and on the server leaving the final
assistant turn open. The fit was changed to read the server's ids, and the check to compare those ids and to close
that turn; both changes were written into the frozen file. No other gate was changed after its data had been
seen. The check that the 66-expert file
with the restored experts switched off reproduces \C{} passed narrowly: relative RMS \LGDisabledRelRms{} against
the \C{}-vs-\C{} ceiling of \LGDisabledRelRmsFloor, and greedy agreement \LGDisabledGreedy\% against the floor of
\LGDisabledGreedyFloor\%. The adapter fixture at every expert layer gave a relative error of \LGFixtureRelRms{}
against a tolerance of \LGFixtureTol.

\paragraph{Third model: the unpruned reference.} The run of the unpruned model on the final problems had its own
file, frozen after the guards and before its first request. It fixed the slot count by a memory measurement, the
settings (those of the baseline's final run, with only the model file changed) and the estimators, and it has no
decision rule.

\paragraph{Third model: compute.} The confirmation and final stages used \LGGpuHours{} hours of generation and
grading, the regression guards \LGGuardGpuHours{} hours, and the unpruned reference \LGRefUWall{} hours.

\section{Development comparison of \A{} and \B{} without retry}
\label{app:dev}

\begin{center}
\begin{minipage}{\linewidth}
\centering
\setlength{\abovecaptionskip}{0pt}
\captionof{table}{Development problems (n = \DevN), one draw, 32K limit, no retry.}
\small
\begin{tabular}{lccc}
\toprule
Arm & Solved & Hit the limit & Difference to \Sarm{} (pp) \\
\midrule
\Sarm & \DevSPass & \DevSCapped & \\
\A{}+\Sarm & \DevASPass & \DevASCapped & \DevASminusSDiff{} \ci{\DevASminusSLo}{\DevASminusSHi} \\
\B{}+\Sarm & \DevBSPass & \DevBSCapped & \DevBSminusSDiff{} \ci{\DevBSminusSLo}{\DevBSminusSHi} \\
\bottomrule
\end{tabular}
\end{minipage}
\end{center}

\B{}+\Sarm{} against \A{}+\Sarm{} is \DevBSminusASDiff\pp{}
\ci{\DevBSminusASLo}{\DevBSminusASHi}. The number of replies hitting the limit is about
the same in all three arms, which is the first sign that the weight methods do not act
mainly on loops.

\end{document}

%% file: numbers.tex
\newcommand{\LoopN}{120}
\newcommand{\LoopCPass}{71}
\newcommand{\LoopCCapped}{44}
\newcommand{\LoopCCappedLoop}{44}
\newcommand{\LoopUtwoPass}{107}
\newcommand{\LoopUtwoCapped}{11}
\newcommand{\LoopUtwoCappedLoop}{0}
\newcommand{\LoopUfourPass}{110}
\newcommand{\LoopUfourCapped}{8}
\newcommand{\LoopUfourCappedLoop}{0}
\newcommand{\LoopPruneDiff}{+36.7}
\newcommand{\LoopPruneLo}{+28.3}
\newcommand{\LoopPruneHi}{+45.8}
\newcommand{\LoopPassUtwoCDiff}{+30.0}
\newcommand{\LoopPassUtwoCLo}{+21.7}
\newcommand{\LoopPassUtwoCHi}{+38.3}
\newcommand{\PracCPass}{71}
\newcommand{\PracCCapped}{44}
\newcommand{\PracSPass}{86}
\newcommand{\PracSCapped}{25}
\newcommand{\PracWPass}{81}

\newcommand{\WN}{124}
\newcommand{\WTestWPass}{68}
\newcommand{\WTestWCapped}{47}

\newcommand{\WTestCPass}{74}
\newcommand{\WTestCCapped}{45}
\newcommand{\WTestCCappedPassed}{0}
\newcommand{\WTestSPass}{80}
\newcommand{\WTestSCapped}{32}

\newcommand{\WOneDiff}{$-$4.8}
\newcommand{\WOneLo}{$-$12.9}
\newcommand{\WOneHi}{+3.2}

\newcommand{\WSminusCDiff}{+4.8}
\newcommand{\WSminusCLo}{$-$2.4}
\newcommand{\WSminusCHi}{+12.1}

\newcommand{\FitARedMedian}{21.8}
\newcommand{\FitARedMin}{14.2}
\newcommand{\FitARedMax}{76.0}
\newcommand{\FitAHours}{5.2}
\newcommand{\FitBRedMedian}{22.8}
\newcommand{\FitBRedMin}{14.5}
\newcommand{\FitBRedMax}{75.8}
\newcommand{\FitBHours}{5.7}

\newcommand{\AdapterBGiB}{1.16}
\newcommand{\FileCGiB}{11.62}
\newcommand{\FileBGiB}{11.88}
\newcommand{\DevSPass}{84}
\newcommand{\DevSCapped}{26}
\newcommand{\DevASPass}{88}
\newcommand{\DevASCapped}{23}
\newcommand{\DevBSPass}{93}
\newcommand{\DevBSCapped}{25}
\newcommand{\DevBSminusSDiff}{+7.5}
\newcommand{\DevBSminusSLo}{+0.8}
\newcommand{\DevBSminusSHi}{+15.0}

\newcommand{\DevASminusSDiff}{+3.3}
\newcommand{\DevASminusSLo}{$-$5.0}
\newcommand{\DevASminusSHi}{+10.8}

\newcommand{\DevBSminusASDiff}{+4.2}
\newcommand{\DevBSminusASLo}{$-$2.5}
\newcommand{\DevBSminusASHi}{+11.7}

\newcommand{\DevN}{120}
\newcommand{\RetrySRPass}{88}

\newcommand{\RetryBSRPass}{101}

\newcommand{\RetryDiff}{+10.8}
\newcommand{\RetryLo}{+3.3}
\newcommand{\RetryHi}{+18.3}
\newcommand{\RetryGained}{18}
\newcommand{\RetryLost}{5}
\newcommand{\RetryP}{0.011}
\newcommand{\TuneCappedStopped}{72}
\newcommand{\TuneCapped}{74}
\newcommand{\TuneFinishedPassedStopped}{2}
\newcommand{\TuneStopMedian}{6{,}003}
\newcommand{\ConfSRonePass}{55}
\newcommand{\ConfBSRonePass}{58}
\newcommand{\ConfSRtwoPass}{52}
\newcommand{\ConfBSRtwoPass}{57}
\newcommand{\ConfN}{64}
\newcommand{\ConfOneDiff}{+4.7}
\newcommand{\ConfOneLo}{$-$1.6}
\newcommand{\ConfOneHi}{+12.5}
\newcommand{\ConfOneGained}{4}
\newcommand{\ConfOneLost}{1}
\newcommand{\ConfOneP}{0.375}
\newcommand{\ConfTwoDiff}{+7.8}
\newcommand{\ConfTwoLo}{$-$1.6}
\newcommand{\ConfTwoHi}{+17.2}
\newcommand{\ConfTwoGained}{7}
\newcommand{\ConfTwoLost}{2}
\newcommand{\ConfTwoP}{0.180}
\newcommand{\ConfPooledDiff}{+6.2}
\newcommand{\ConfPooledLo}{+0.8}
\newcommand{\ConfPooledHi}{+13.3}
\newcommand{\FinalN}{130}
\newcommand{\FinalDiff}{+9.2}
\newcommand{\FinalLo}{+3.1}
\newcommand{\FinalHi}{+15.4}
\newcommand{\FinalGained}{15}
\newcommand{\FinalLost}{3}
\newcommand{\FinalP}{0.007}
\newcommand{\FinalDecision}{leads}
\newcommand{\FinalFirstDrawDiff}{+8.5}
\newcommand{\FinalFirstDrawLo}{+0.8}
\newcommand{\FinalFirstDrawHi}{+16.2}
\newcommand{\FinalSRPass}{104}
\newcommand{\FinalSRPassPct}{80.0}
\newcommand{\FinalSRWithinOne}{89}

\newcommand{\FinalSRDraws}{175}
\newcommand{\FinalSRTokens}{707}
\newcommand{\FinalSRWall}{2.73}
\newcommand{\FinalSRFailWrong}{16}
\newcommand{\FinalSRFailLoop}{9}

\newcommand{\FinalSREasy}{60}
\newcommand{\FinalEasyN}{63}
\newcommand{\FinalSRMedium}{44}
\newcommand{\FinalMediumN}{67}

\newcommand{\FinalBSRPass}{116}
\newcommand{\FinalBSRPassPct}{89.2}
\newcommand{\FinalBSRWithinOne}{100}

\newcommand{\FinalBSRDraws}{162}
\newcommand{\FinalBSRTokens}{791}
\newcommand{\FinalBSRWall}{4.63}
\newcommand{\FinalBSRFailWrong}{8}
\newcommand{\FinalBSRFailLoop}{4}

\newcommand{\FinalBSREasy}{62}
\newcommand{\FinalBSRMedium}{54}

\newcommand{\FinalWallRatio}{1.7}
\newcommand{\GpuHours}{14.5}
\newcommand{\PeakGiB}{81.6}
\newcommand{\GuardVerdict}{hold}
\newcommand{\GuardHEN}{164}
\newcommand{\GuardHESR}{144}
\newcommand{\GuardHEBSR}{145}
\newcommand{\GuardHEDiff}{+0.6}
\newcommand{\GuardHELo}{$-$4.3}
\newcommand{\GuardHEHi}{+5.5}

\newcommand{\GuardMBN}{378}
\newcommand{\GuardMBSR}{276}
\newcommand{\GuardMBBSR}{269}
\newcommand{\GuardMBDiff}{$-$1.9}
\newcommand{\GuardMBLo}{$-$5.6}
\newcommand{\GuardMBHi}{+1.6}
\newcommand{\GuardMBGained}{20}
\newcommand{\GuardMBLost}{27}

\newcommand{\FNFileCGiB}{45.50}
\newcommand{\FNFileBGiB}{45.72}
\newcommand{\FNAdapterBGiB}{1.79}
\newcommand{\FNFileUGiB}{87.25}
\newcommand{\FNParityShortLog}{0.0042}
\newcommand{\FNParityShortTol}{0.03}
\newcommand{\FNParityLongLog}{0.0135}

\newcommand{\FNFitRedMedian}{30.1}
\newcommand{\FNFitRedMin}{17.9}
\newcommand{\FNFitRedMax}{55.0}
\newcommand{\FNDevN}{120}
\newcommand{\FNDevCLoop}{64}
\newcommand{\FNDisabledGreedyMin}{99}
\newcommand{\FNDisabledGreedy}{96.1}
\newcommand{\FNDisabledFloorLo}{95.3}
\newcommand{\FNDisabledFloorHi}{98.4}
\newcommand{\FNFixtureRelRms}{0.084}
\newcommand{\FNFixtureTol}{0.05}
\newcommand{\FNFixtureResidualCorr}{0.03}
\newcommand{\FNFixtureAllRelRms}{0.042}
\newcommand{\FNTokShared}{248{,}070}
\newcommand{\FNTokExtra}{250}
\newcommand{\FNConfSRonePass}{34}
\newcommand{\FNConfBSRonePass}{56}
\newcommand{\FNConfSRtwoPass}{37}
\newcommand{\FNConfBSRtwoPass}{56}
\newcommand{\FNConfOneDiff}{+34.4}
\newcommand{\FNConfOneLo}{+23.4}
\newcommand{\FNConfOneHi}{+45.3}
\newcommand{\FNConfOneGained}{22}
\newcommand{\FNConfOneLost}{0}
\newcommand{\FNConfOneP}{<0.001}
\newcommand{\FNConfTwoDiff}{+29.7}
\newcommand{\FNConfTwoLo}{+18.8}
\newcommand{\FNConfTwoHi}{+40.6}
\newcommand{\FNConfTwoGained}{19}
\newcommand{\FNConfTwoLost}{0}
\newcommand{\FNConfTwoP}{<0.001}
\newcommand{\FNConfN}{64}
\newcommand{\FNConfPooledDiff}{+32.0}
\newcommand{\FNConfPooledLo}{+22.7}
\newcommand{\FNConfPooledHi}{+41.4}
\newcommand{\FNFinalN}{130}
\newcommand{\FNFinalDiff}{+35.4}
\newcommand{\FNFinalLo}{+26.2}
\newcommand{\FNFinalHi}{+44.6}
\newcommand{\FNFinalGained}{49}
\newcommand{\FNFinalLost}{3}
\newcommand{\FNFinalP}{<0.001}
\newcommand{\FNFinalDecision}{leads}
\newcommand{\FNFinalFirstDrawDiff}{+30.0}
\newcommand{\FNFinalFirstDrawLo}{+21.5}
\newcommand{\FNFinalFirstDrawHi}{+38.5}
\newcommand{\FNFinalSRPass}{63}
\newcommand{\FNFinalSRDraws}{201}

\newcommand{\FNFinalSRTokens}{1{,}052}
\newcommand{\FNFinalSRFailWrong}{40}
\newcommand{\FNFinalSRFailLoop}{26}
\newcommand{\FNFinalSREasy}{45}
\newcommand{\FNFinalEasyN}{63}
\newcommand{\FNFinalSRMedium}{18}
\newcommand{\FNFinalMediumN}{67}
\newcommand{\FNFinalBSRPass}{109}
\newcommand{\FNFinalBSRDraws}{176}

\newcommand{\FNFinalBSRTokens}{1{,}649}
\newcommand{\FNFinalBSRFailWrong}{10}
\newcommand{\FNFinalBSRFailLoop}{9}
\newcommand{\FNFinalBSREasy}{61}
\newcommand{\FNFinalBSRMedium}{48}
\newcommand{\FNFinalWallRatio}{2.5}
\newcommand{\FNGuardVerdict}{hold}
\newcommand{\FNGuardGpuHours}{16}
\newcommand{\FNGuardHEN}{164}
\newcommand{\FNGuardHESR}{138}
\newcommand{\FNGuardHEBSR}{147}
\newcommand{\FNGuardHEDiff}{+5.5}
\newcommand{\FNGuardHELo}{+0.6}
\newcommand{\FNGuardHEHi}{+10.4}

\newcommand{\FNGuardMBN}{378}
\newcommand{\FNGuardMBSR}{282}
\newcommand{\FNGuardMBBSR}{285}
\newcommand{\FNGuardMBDiff}{+0.8}
\newcommand{\FNGuardMBLo}{$-$2.6}
\newcommand{\FNGuardMBHi}{+4.2}

\newcommand{\FNGpuHours}{27.5}
\newcommand{\FNBudgetHours}{48}
\newcommand{\FNPeakGiB}{88.6}
\newcommand{\LGFileUGiB}{53.61}
\newcommand{\LGYarnGGUF}{32}
\newcommand{\LGSlidingWindow}{512}
\newcommand{\LGFileCGiB}{16.31}
\newcommand{\LGParityUShortLog}{$-$0.0034}
\newcommand{\LGParityULongLog}{0.0015}
\newcommand{\LGParityCShortLog}{0.0073}
\newcommand{\LGParityCLongLog}{$-$0.0030}
\newcommand{\LGParityTol}{0.03}
\newcommand{\LGFitRedMedian}{44.6}
\newcommand{\LGFitRedMin}{14.2}
\newcommand{\LGFitRedMax}{92.0}
\newcommand{\LGDisabledRelRms}{0.00796}
\newcommand{\LGDisabledRelRmsFloor}{0.00799}
\newcommand{\LGDisabledGreedy}{97.7}
\newcommand{\LGDisabledGreedyFloor}{94.5}
\newcommand{\LGFixtureRelRms}{0.044}
\newcommand{\LGFixtureTol}{0.05}
\newcommand{\LGFitTextBeforeLoop}{15.1}
\newcommand{\LGFitTextBeforeLoopOrnith}{21.6}
\newcommand{\LGFitTextBeforeLoopFlashnext}{47.6}
\newcommand{\LGTokDocs}{1{,}274}
\newcommand{\LGTokDocsCarriageReturn}{267}
\newcommand{\LGYarnConfig}{128}
\newcommand{\LGFullAttnLayers}{12}
\newcommand{\LGLayers}{48}
\newcommand{\LGScreenN}{120}
\newcommand{\LGScreenUPass}{107}
\newcommand{\LGScreenCPass}{34}
\newcommand{\LGConfSRonePass}{22}
\newcommand{\LGConfBSRonePass}{28}
\newcommand{\LGConfSRtwoPass}{17}
\newcommand{\LGConfBSRtwoPass}{27}
\newcommand{\LGConfOneDiff}{+9.4}
\newcommand{\LGConfOneLo}{$-$4.7}
\newcommand{\LGConfOneHi}{+23.4}
\newcommand{\LGConfOneGained}{13}
\newcommand{\LGConfOneLost}{7}
\newcommand{\LGConfOneP}{0.263}
\newcommand{\LGConfTwoDiff}{+15.6}
\newcommand{\LGConfTwoLo}{+3.1}
\newcommand{\LGConfTwoHi}{+28.1}
\newcommand{\LGConfTwoGained}{14}
\newcommand{\LGConfTwoLost}{4}
\newcommand{\LGConfTwoP}{0.031}
\newcommand{\LGConfN}{64}
\newcommand{\LGConfPooledDiff}{+12.5}
\newcommand{\LGConfPooledLo}{+2.3}
\newcommand{\LGConfPooledHi}{+22.7}
\newcommand{\LGFinalN}{130}
\newcommand{\LGFinalDiff}{+16.2}
\newcommand{\LGFinalLo}{+6.9}
\newcommand{\LGFinalHi}{+25.4}
\newcommand{\LGFinalGained}{31}
\newcommand{\LGFinalLost}{10}
\newcommand{\LGFinalP}{0.002}
\newcommand{\LGFinalDecision}{leads}
\newcommand{\LGFinalFirstDrawDiff}{+13.9}
\newcommand{\LGFinalFirstDrawLo}{+4.6}
\newcommand{\LGFinalFirstDrawHi}{+23.8}
\newcommand{\LGFinalSRPass}{36}
\newcommand{\LGFinalSRDraws}{293}
\newcommand{\LGFinalSRLoopDraws}{220}
\newcommand{\LGFinalSRTokens}{1{,}280}
\newcommand{\LGFinalSRFailWrong}{39}
\newcommand{\LGFinalSRFailLoop}{55}
\newcommand{\LGFinalSREasy}{27}
\newcommand{\LGFinalEasyN}{63}
\newcommand{\LGFinalSRMedium}{9}
\newcommand{\LGFinalMediumN}{67}
\newcommand{\LGFinalBSRPass}{57}
\newcommand{\LGFinalBSRDraws}{214}

\newcommand{\LGFinalBSRTokens}{860}
\newcommand{\LGFinalBSRFailWrong}{49}
\newcommand{\LGFinalBSRFailLoop}{24}
\newcommand{\LGFinalBSREasy}{38}
\newcommand{\LGFinalBSRMedium}{19}
\newcommand{\LGFinalWallRatio}{1.1}
\newcommand{\LGGuardVerdict}{hold}
\newcommand{\LGGuardGpuHours}{12.5}
\newcommand{\LGGuardHEN}{164}
\newcommand{\LGGuardHESR}{50}
\newcommand{\LGGuardHEBSR}{86}
\newcommand{\LGGuardHEDiff}{+22.0}
\newcommand{\LGGuardHELo}{+13.4}
\newcommand{\LGGuardHEHi}{+30.5}

\newcommand{\LGGuardMBN}{378}
\newcommand{\LGGuardMBSR}{145}
\newcommand{\LGGuardMBBSR}{161}
\newcommand{\LGGuardMBDiff}{+4.2}
\newcommand{\LGGuardMBLo}{$-$1.3}
\newcommand{\LGGuardMBHi}{+10.1}

\newcommand{\LGGpuHours}{14.5}
\newcommand{\LGBudgetHours}{37.9}
\newcommand{\LGPeakGiB}{57.4}
\newcommand{\LGFinalSREmptyThink}{183}

\newcommand{\LGRefUPass}{119}
\newcommand{\LGRefBelowU}{47.7}
\newcommand{\LGRefBelowULo}{39.2}
\newcommand{\LGRefBelowUHi}{56.9}
\newcommand{\LGRefShare}{25.3}
\newcommand{\LGRefShareLo}{11.4}
\newcommand{\LGRefShareHi}{38.1}
\newcommand{\LGRefGap}{83}
\newcommand{\LGRefClosed}{21}
\newcommand{\LGRefUDraws}{130}
\newcommand{\LGRefUWall}{1.98}
\newcommand{\LGRefPeakGiB}{94.3}
\newif\ifguardsdone\guardsdonetrue